\documentclass[11pt, a4paper, onecolumn, copyright]{handshaketemplate}

\usepackage[
  backend=biber,
  style=authoryear,
  natbib=true,
  maxcitenames=2,
  maxbibnames=99,
  uniquelist=false,
  uniquename=false,
  dashed=false,
  mincrossrefs=99,
]{biblatex}
\DeclareDelimFormat{nameyeardelim}{\addcomma\space}
\DeclareDelimFormat{multinamedelim}{\addsemicolon\space}
\DeclareDelimFormat{finalnamedelim}{\addspace\&\space}
\DeclareDelimFormat{multicitedelim}{\addsemicolon\space}

\AtEveryBibitem{\clearfield{urldate}\clearfield{urlyear}\clearfield{urlmonth}\clearfield{urlday}}

\newcommand{\ishandshake}{}

\newif\ifvariantneurips
\newif\ifvarianthandshake
\newif\ifvariantworkshop
\newif\ifvariantrlevalcais
\newif\ifvariantkdd
\newif\ifvariantposter

\newcommand{\onlykdd}[1]{\ifvariantkdd#1\fi}

\newcommand{\skipkdd}[1]{\ifvariantkdd\else#1\fi}

\varianthandshaketrue

\usepackage{bm}
\usepackage[font=small,labelfont=rm]{subcaption}
\usepackage{graphicx}
\usepackage[dvipsnames]{xcolor}

\usepackage{tikz}
\usetikzlibrary{arrows.meta, calc, positioning, shapes.geometric, decorations.pathreplacing, patterns}
\usepackage{pgfplots}
\usepackage{pgfplotstable}
\usepgfplotslibrary{groupplots, fillbetween, statistics}
\pgfplotsset{compat=1.18}

\definecolor{bluecolorsbase}     {HTML}{0072B2} %
\definecolor{orangecolorsbase}   {HTML}{E69F00} %
\definecolor{greencolorsbase}    {HTML}{009E73} %
\definecolor{purplecolorsbase}   {HTML}{CC79A7} %
\definecolor{redcolorsbase}      {HTML}{D55E00} %
\definecolor{yellowcolorsbase}   {HTML}{F0E442} %
\definecolor{skybluecolorsbase}  {HTML}{56B4E9} %
\definecolor{graycolorsbase}     {HTML}{4D4D4D} %

\newcommand{\makeshades}[2]{%
    \colorlet{#1light2}{#2!25!white}%
    \colorlet{#1light1}{#2!55!white}%
    \colorlet{#1main}  {#2}%
    \colorlet{#1dark1} {#2!75!black}%
    \colorlet{#1dark2} {#2!50!black}%
}

\makeshades{bluecolors}    {bluecolorsbase}
\makeshades{orangecolors}  {orangecolorsbase}
\makeshades{greencolors}   {greencolorsbase}
\makeshades{purplecolors}  {purplecolorsbase}
\makeshades{redcolors}     {redcolorsbase}
\makeshades{yellowcolors}  {yellowcolorsbase}
\makeshades{skybluecolors} {skybluecolorsbase}
\makeshades{graycolors}    {graycolorsbase}

\newcommand{\makeverifieralias}[2]{%
    \colorlet{#1Light2}{#2light2}%
    \colorlet{#1Light1}{#2light1}%
    \colorlet{#1Main}  {#2main}%
    \colorlet{#1Dark1} {#2dark1}%
    \colorlet{#1Dark2} {#2dark2}%
}

\makeverifieralias{gandalf}    {bluecolors}
\makeverifieralias{archipelago}{orangecolors}
\makeverifieralias{aaj}        {greencolors}
\makeverifieralias{autorubric}        {redcolors}
\makeverifieralias{rewardkit}  {purplecolors}

\usepackage{standalone}
\usepackage{tcolorbox}
\tcbuselibrary{breakable}
\usepackage{placeins}
\usepackage{fvextra}
\usepackage{csquotes}
\usepackage{xspace}
\usepackage{pifont}
\newcommand{\cmark}{\ding{51}}
\newcommand{\xmark}{\ding{55}}

\ifdefined\ishandshake
  
\else
  
\fi

\makeatletter
\newcommand{\@btbcite}[1]{%
  \ifanonymized
    #1{anon:bankertoolbench}%
  \else
    #1{lau:bankertoolbench}%
  \fi
  \xspace
}
\newcommand{\citepbtb}{\@btbcite{\citep}}
\newcommand{\citetbtb}{\@btbcite{\citet}}
\makeatother

\newcommand{\gandalfurl}{%
  \ifanonymized
    \url{https://anonymous.4open.science/r/gandalf-the-grader-0E96}%
  \else
    \url{https://github.com/Handshake-AI-Research/gandalf-the-grader}%
  \fi
}

\newcommand{\bvbdataurl}{\url{https://huggingface.co/datasets/handshake-ai-research/BankerVerifierBench/tree/v1.0}}
\newcommand{\bvbcodeurl}{\url{https://github.com/Handshake-AI-Research/bankerverifierbench/tree/v1.0}}

\newcommand{\gandalf}{Gandalf\xspace}
\newcommand{\benchmark}{BankerVerifierBench\xspace}
\newcommand{\benchmarkabbrev}{BVB\xspace}

\crefformat{section}{\S#2#1#3}
\crefformat{subsection}{\S#2#1#3}
\crefformat{subsubsection}{\S#2#1#3}

\usepackage{xurl}
\AtBeginEnvironment{tabular}{\footnotesize}
\RecustomVerbatimEnvironment{verbatim}{Verbatim}{%
  breaklines=true,breakanywhere=true,breaksymbolleft={},breaksymbolright={}}
\newif\ifanonymized

\definecolor{darkblue}{rgb}{0, 0, 0.5}
\hypersetup{colorlinks=true, citecolor=darkblue, linkcolor=darkblue, urlcolor=darkblue}

\newcommand{\cornerlogo}[1]{%
  \begin{tikzpicture}[remember picture,overlay]
    \node[anchor=north west, xshift=2.0cm, yshift=-1.8cm] at (current page.north west)
      {#1};
    \draw[line width=0.3pt]
      ($(current page.north west)+(2.0cm,-2.7cm)$) -- ($(current page.north east)+(-2.0cm,-2.7cm)$);
  \end{tikzpicture}%
}

\title{Verifying Agents in Rubric-Graded Environments}

\renewcommand{\today}{}

\makeatletter
\renewcommand\AB@authnote[1]{\textsuperscript{#1}}
\renewcommand\AB@affilnote[1]{\textsuperscript{#1}}
\makeatother

\author[1]{Markus~D\"ucker}
\author[1]{Vaibhav~Kumar}
\author[1]{Yi~Liu}
\author[1]{Ronak~Chaudhary}
\author[1,2]{Andreas~Plesner}
\author[1]{Francisco~Guzm\'an}
\author[1]{Anish~Athalye}
\affil[1]{Handshake AI Research}
\affil[2]{ETH Z\"urich}

\begin{abstract}
As AI agents take on increasingly open-ended and complex tasks, verifying their outputs has proven to be difficult. 
Rubric-graded agent environments, where a verifier judges multiple natural-language criteria against the deliverables and environment state changes produced by
the agent, have emerged as a popular paradigm. 
We conduct the first systematic study of verifiers for rubric-graded agent environments.
First, we create \benchmark~(\benchmarkabbrev), a meta-evaluation dataset of $2{,}464$ human-judged rubric criteria across $24$ investment-banking tasks. 
Next, we develop a methodology that derives from any rubric corpus the capabilities a verifier should possess, yielding a nine-capability taxonomy that we distill into three general design principles---reactive verification, environment alignment, and domain guidance---which we implement in \gandalf, an open-source verifier. Finally, we evaluate \gandalf against three existing verifiers on \benchmarkabbrev. \gandalf matches or exceeds every baseline's F1 at lower cost: with Gemini~3~Flash it matches the strongest baseline's F1 (0.864 vs.\ 0.852) at one-fifth of its cost, and with GPT-5.4 models it improves F1 by over 0.12 at lower cost. \gandalf also generalizes to OpenClaw, a structurally different personal-productivity environment, where it again achieves the highest F1 (0.955 vs.\ 0.934 for the best baseline). We release \benchmarkabbrev, \gandalf, and the code and data to reproduce our experiments.
\end{abstract}

\begin{document}

\cornerlogo{\includegraphics[scale=0.27]{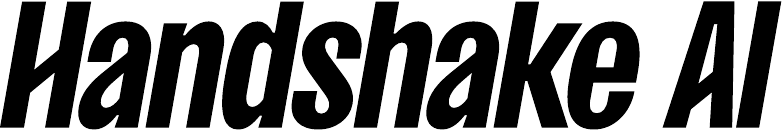}}

\maketitle

\section{Introduction}

Modern AI agents are systems that use large language models (LLMs) to interact with tools and environments in pursuit of a goal~\citep{gao:pal,schick:toolformer,yao:react}. As these agents have grown increasingly capable and reliable, so has the need to measure and optimize their performance. Measurement and optimization go hand-in-hand: agents are evaluated within \emph{environments}, and these same environments can be used for reinforcement learning (RL) to train improved models~\citep{luo:agent-lightning}. While some agentic benchmarks such as GDPval~\citep{patwardhan:gdpval} are human-graded using pairwise preferences, enabling comparison between agents on leaderboards, the vast majority strive for automated pointwise grading, facilitating both benchmarking and reinforcement learning by using the grade as the reward signal. Both uses depend on a verifier whose grades can be trusted.

Early efforts, encouraged by the success of reinforcement learning with verifiable rewards~(RLVR)~\citep{shao:grpo,guo:deepseek-r1}, focused on \emph{verifiable domains} where deterministic automated verification is possible. For example, SWE-bench~\citep{jimenez:swe-bench} uses unit tests to verify software engineering, HotpotQA~\citep{yang:hotpotqa} uses exact-match scoring for multi-hop question-answering, and WebArena~\citep{zhou:webarena} uses programmatic checks for browser use. Such environments trade off verifiability against scope, open-endedness, and realism.

Recently, agent evaluation has expanded into \emph{semi-verifiable domains} where judgment is required. Benchmarks such as BankerToolBench~\citepbtb, APEX-Agents~\citep{vidgen:apex-agents}, CoreCraft~\citep{mehta:corecraft}, and ClawBench~\citep{zhang:clawbench} use natural-language \emph{rubrics}---fine-grained binary criteria that are objectively gradable---as a technique to define and capture correctness~\citep{viswanathan:rlcf,he:rifl,gunjal:rar}.

Verifying agent rollouts in such environments is a challenging task in its own right. Agents may produce multi-file deliverables (e.g., an Excel spreadsheet and a PowerPoint presentation) and invoke tools with side effects (e.g., sending an email). Verification requires checking diverse properties such as whether an email was sent, whether an Excel model's formulas propagate correctly across tabs, and whether a PowerPoint's charts are internally consistent with the underlying data.

Verifiers for rubric-graded environments span a broad design space, ranging from using an LLM-as-a-judge on the trajectory~\citep{mehta:corecraft} to making the verifier itself agentic: an \emph{agent-as-a-judge}~\citep{zhuge:agent-as-a-judge,you:agent-as-a-judge-survey}. Agentic judges themselves range from workflows (e.g., Archipelago~\citep{vidgen:apex-agents}) to fully reactive agents~\citep{yao:react} (e.g., ClawBench~\citep{zhang:clawbench}). Despite growing interest from both research and industry, existing benchmarks treat the verifier as a subproblem---a means to grade rollouts or supply training rewards---rather than as an object of study. This gap, combined with the broad design space, motivates studying these verifiers in their own right.

To the best of our knowledge, we conduct the first systematic study of verifiers for agents in rubric-graded environments. To enable this, we produce a dataset of human annotations, analogous to prior work on meta-evaluation of LLM judging and rubric-based evaluation~\citep{tan:judgebench,pan:rubriceval}. The dataset consists of agent trajectories and final environment states (containing file deliverables), along with rubric criteria and human-produced judgments, across a number of tasks in the BankerToolBench~(BTB)~\citepbtb agent environment. We call this verifier evaluation dataset \benchmark~(\benchmarkabbrev). \benchmarkabbrev consists of $2{,}464$ judgment instances across $24$ tasks, each judged independently by two domain experts and adjudicated by a senior reviewer. We publicly release \benchmarkabbrev together with the code to reproduce our experiments, to facilitate future work on evaluating and improving verifiers for rubric-graded environments.

We develop a methodology to analyze rubrics and identify the raw \emph{capabilities} a verifier needs to judge the criteria in a dataset. Applying this methodology to \benchmarkabbrev, we identify nine capabilities that lead us to three general design principles for verifiers in rubric-graded environments. We implement these principles in \emph{\gandalf}\footnote{\gandalfurl{}\hspace{0.5em}(v1.0.0)}, a new open-source verifier for agent environments.

Our evaluation against three existing verifiers shows that \gandalf offers the best accuracy--cost tradeoff on \benchmarkabbrev (\Cref{sec:evaluation}): for every baseline configuration, a cheaper \gandalf configuration is at least as accurate. Across all nine capabilities on \benchmarkabbrev, no baseline is reliably more accurate than \gandalf (\Cref{sec:discussion-capability}). On OpenClaw, an internal dataset from a structurally different personal-productivity environment, \gandalf again achieves the highest accuracy (\Cref{sec:discussion-generalize}). No other verifier is near the top on both datasets: Archipelago comes close on \benchmarkabbrev but falls behind on OpenClaw, and AutoRubric does the reverse.

\section{Background}
\label{sec:background}
In a \emph{rubric-graded agent environment}, an agent receives a task
and acts within a structured environment (e.g., a filesystem, a set of
tools, and access to external data) to produce a deliverable, which
may consist of one or more files together with changes to the
environment. A verifier is then asked to assign a binary met/unmet
judgment to each criterion in the task's natural-language rubric.
This contrasts with deterministically graded agent benchmarks, where
criteria reduce to executable checks. Rubric-graded criteria are open-ended
natural-language descriptions whose judgment often requires
inspection of the deliverable, domain knowledge, and qualitative
assessment. We adopt BankerToolBench (BTB) as the basis for our study and
select three representative verifiers as objects of study.
 
\paragraph{BankerToolBench (BTB) environment.}
We use version 1.1 of BTB~\citepbtb, a recently released benchmark of
$100$ end-to-end analytical workflows performed by junior investment
bankers, including discounted cash-flow analyses, leveraged-buyout
models, trading comparables, and merger models. Each task
is long-horizon, taking a banker $5$ hours on average.
The agent receives a task description, a data room of pre-loaded
files, and a fixed set of tools: a market-data platform, an SEC EDGAR
API, and a company brand-assets API for logos, alongside a filesystem and
pre-installed libraries such as \texttt{openpyxl} and
\texttt{python-pptx}. It must produce multi-file deliverables in
standard office formats (\texttt{.xlsx}, \texttt{.pptx},
\texttt{.docx}, \texttt{.pdf}), which are graded against an
expert-authored rubric of binary, weighted criteria. BTB rollouts
are a demanding setting for verification: judging a single criterion
may require joint reasoning across multiple heterogeneous artifacts.
 
\paragraph{Verifiers for rubric-graded environments.}
We study three open-source verifiers that engage with the deliverable to
different degrees. \emph{AutoRubric}~\citep{rao:autorubric} grades
each criterion in an independent LLM call over the full agent
trajectory, including recorded tool calls with their arguments and outputs,
but has no direct access to the deliverable files.
\emph{Archipelago}~\citep{vidgen:apex-agents}, released with
APEX-Agents, captures a filesystem diff between initial and final
environment states, extracts the content of changed documents via the
Reducto document extraction service, and grades each criterion in an independent LLM
call over the diff and extracted content without interacting
with the deliverable directly.
\emph{Agent-as-a-Judge}~\citep{zhuge:agent-as-a-judge} employs
a workflow-based agent as the evaluator, planning and executing steps
such as locating and reading files to reach a verdict.

\section{\benchmark: Meta-evaluation for rubric-graded environments}
\label{sec:bvb}
We introduce BankerVerifierBench (BVB), the first meta-evaluation
dataset for verifiers in rubric-graded agent environments. BVB
extends prior meta-evaluation work for LLM judges
\citep{tan:judgebench,pan:rubriceval} and agent trajectories
\citep{lu:agentrewardbench} to settings where criteria require
inspecting multi-file binary deliverables and tool-mediated
environment state rather than text alone. The dataset consists of
$24$ tasks drawn from the BTB task distribution, spanning the product
groups and task types described in \Cref{sec:background}.
We randomly sampled $24$ of BTB’s $100$ tasks, choosing the largest sample our annotation budget allowed.
Each task ships with an expert-authored rubric of binary criteria, weighted by importance and tagged with one of six categories; rubrics range from $67$ to $128$ criteria per task (median $104$), totaling $2{,}464$ criteria across the corpus.
Each task is paired with one selected agent
rollout, generated using Gemini 2.5 Pro or GLM-5 (12 tasks each, to avoid tying the labels to one agent's failure modes) with OpenCode 1.17.9 as the agent harness. Rollouts average 25.5 tool calls (median 24, range 3--52), and generating the 24 released rollouts cost \$14.36 in model API usage (mean \$0.60, median \$0.29 per rollout). Costs are as charged via OpenRouter and reflect prompt-caching discounts (\Cref{app:rollouts}). Each rollout includes the complete trajectory (messages, tool calls, tool responses) and the final workspace, including all deliverable files in their native binary formats.
Two practicing investment bankers, compensated at or above prevailing US market rates, independently judged each criterion in the native deliverables with a binary met/unmet verdict, with a raw inter-annotator agreement of $92.1\%$ and a Cohen's $\kappa$~\citep{cohen:kappa} of $0.82$. A senior reviewer adjudicated the labels, resolving disagreements and correcting judgements where necessary. The final adjudicated labels serve as our evaluation ground truth; $66.3\%$ of criteria are met.

\paragraph{Release.}
We release \benchmarkabbrev under a CC-BY-4.0 license at \bvbdataurl, with a dataset card documenting its contents, provenance, and derivation from BankerToolBench~\citepbtb, and the code to reproduce our \benchmarkabbrev experiments under Apache-2.0 at \bvbcodeurl. The OpenClaw evaluation set (\Cref{sec:discussion-generalize}) is internal and not released; we document its methods and results, but they cannot be independently reproduced without it.

\section{Deriving requirements for verifiers from data}
\label{sec:taxonomy}
\newcommand{\capcriteria}{2{,}464}
\newcommand{\capn}[1]{\csname capn:#1\endcsname}
\newcommand{\capshare}[1]{\csname capshare:#1\endcsname}
\expandafter\def\csname capn:quantitative\endcsname{522}
\expandafter\def\csname capshare:quantitative\endcsname{21.2}
\expandafter\def\csname capn:visual\endcsname{406}
\expandafter\def\csname capshare:visual\endcsname{16.5}
\expandafter\def\csname capn:literal\endcsname{399}
\expandafter\def\csname capshare:literal\endcsname{16.2}
\expandafter\def\csname capn:structural\endcsname{322}
\expandafter\def\csname capshare:structural\endcsname{13.1}
\expandafter\def\csname capn:crosscomponent\endcsname{227}
\expandafter\def\csname capshare:crosscomponent\endcsname{9.2}
\expandafter\def\csname capn:labeling\endcsname{223}
\expandafter\def\csname capshare:labeling\endcsname{9.1}
\expandafter\def\csname capn:governance\endcsname{129}
\expandafter\def\csname capshare:governance\endcsname{5.2}
\expandafter\def\csname capn:temporal\endcsname{123}
\expandafter\def\csname capshare:temporal\endcsname{5.0}
\expandafter\def\csname capn:accounting\endcsname{113}
\expandafter\def\csname capshare:accounting\endcsname{4.6}

Tasks in rubric-graded agent environments come paired with a rubric of natural-language criteria. 
Rather than fixing a verifier architecture in advance and determining its capabilities, we start from the rubric and ask what capabilities does a verifier need to have. We describe a general methodology for deriving
verifier requirements from any rubric corpus, then apply its taxonomy to
\benchmarkabbrev. The methodology produces a taxonomy of verifier-side
\emph{capabilities}: each capability names a kind of check the
verifier must perform to judge a criterion correctly. The taxonomy is
verifier-side (labels describe what a verifier must check, not what
the agent must produce) and architecture-agnostic (labels do not name
a specific runtime, library, file format, or data source), so it
applies across verifier types.

\paragraph{Methodology.}
Our pipeline takes a corpus of natural-language rubrics and produces
a taxonomy of capabilities in three stages. \emph{Stage~1} generates
open-ended natural-language tags for each criterion describing what
a verifier needs to do to verify it. \emph{Stage~2} clusters these
tags and merges them into a small set of canonical capabilities.
\emph{Stage~3} re-tags every criterion with one to three of the
canonical capabilities, ranked by relevance. We provide more details about the methodology in
\Cref{app:taxonomy-construction}.
 
\paragraph{Application to \benchmarkabbrev.}
 
We retain the nine capabilities induced from an earlier
\benchmarkabbrev{} rubric corpus and apply them to the released
criteria.\footnote{That corpus was based on BTB~v1.0; for the released
BTB~v1.1 corpus, we kept the definitions
fixed and reran Stage~3 on all $\capcriteria$ criteria, each of which
received a valid tag.}
\Cref{tab:taxonomy} reports their primary-tag distribution.
Across the released corpus, the largest, \emph{audit quantitative metric
calculation}, accounts for $\capshare{quantitative}\%$; the smallest,
\emph{audit accounting logic and convention}, accounts for
$\capshare{accounting}\%$.
 
\begin{table*}[t]
\caption{Retained nine-capability taxonomy applied to the released
\benchmarkabbrev{} corpus. $n$ counts the $\capcriteria$ criteria whose
primary Stage~3 tag is the capability; \% is the corresponding share.}
\label{tab:taxonomy}
\centering
\small
\setlength{\tabcolsep}{6pt}
\renewcommand{\arraystretch}{1.25}
\begin{tabular}{>{\raggedright}p{5.2cm}>{\raggedright\arraybackslash}p{6.0cm}rr}
\toprule
Capability & Description & $n$ & \% \\
\midrule
Audit quantitative metric calculation &
Recompute a derived figure from its inputs and compare against the
deliverable. & \capn{quantitative} & \capshare{quantitative} \\
Inspect visual presentation and style &
Evaluate fonts, colors, alignment, layout, and branding against the
specification. & \capn{visual} & \capshare{visual} \\
Verify literal data point accuracy &
Confirm a constant in the deliverable matches a constant in the brief
or source data. & \capn{literal} & \capshare{literal} \\
Verify structural component inventory &
Confirm the deliverable contains the prescribed sections, files, and
columns. & \capn{structural} & \capshare{structural} \\
Reconcile cross-component data linkage &
Trace a value across two locations in the deliverable (across sheets,
tabs, or files). & \capn{crosscomponent} & \capshare{crosscomponent} \\
Validate document labeling and metadata &
Confirm textual labels (tab names, axis titles, headers) match the
specification verbatim. & \capn{labeling} & \capshare{labeling} \\
Audit governance and strategic narrative &
Confirm required disclosures are present and qualitative claims cohere with the analysis. & \capn{governance} & \capshare{governance} \\
Validate temporal and sequential integrity &
Verify chronological alignment, period sequencing, and stage-boundary
integrity. & \capn{temporal} & \capshare{temporal} \\
Audit accounting logic and convention &
Check signs, identities, and component completeness against domain
conventions. & \capn{accounting} & \capshare{accounting} \\
\bottomrule
\end{tabular}
\end{table*}

In the \benchmarkabbrev setting, these abstract capabilities take
concrete form. \emph{Audit quantitative metric calculation} ranges
over EV/EBITDA multiples and transaction fees; \emph{audit
accounting logic and convention} covers sign rules and balance-sheet
identities; \emph{audit governance and strategic narrative} covers
disclaimers, regulatory boilerplate, and acquisition-rationale claims.
The same capability labels might carry to other domains although with different
instantiations: for example, a clinical workflow would instantiate \emph{audit
accounting logic and convention} as dosage and contraindication
checks against clinical guidelines; a productivity workflow would
instantiate \emph{audit governance and strategic narrative} as
escalation policies and tone norms.

\section{Design principles for verifiers}
\label{sec:design-principles}
 
Reading the taxonomy as a specification surfaces three structural
patterns in the capabilities: (i) the verification path varies across
criteria, (ii) judging requires inspecting artifact-internal structure or environment state
that text serialization erases, and (iii) judging requires domain
knowledge not in the criterion text. These patterns arise from natural-language rubrics over rich agent environments, not specifically from the financial setting of \benchmarkabbrev. We address each with a design principle.
 
\paragraph{Principle 1 (P1): Reactive verification.}
We cannot predict the verification path from the criterion text
alone: two criteria within a single capability can require different
files to open, different tools to invoke, and different orders of
operation, and the number of distinct paths grows with the rubric.
The pattern is intrinsic to natural-language rubrics rather than a
quirk of \benchmarkabbrev: any domain that admits rubrics over
open-ended deliverables will exhibit it. Our first principle is that the verifier should
be a \emph{reactive agent}~\citep{yao:react} that, given a criterion, autonomously
decides which artifacts to open, which tools to invoke, and in what
order, terminating once it reaches a verdict. A workflow verifier fixes these steps when it is designed, while a reactive verifier lets the model choose them for each criterion. Every capability in the taxonomy exhibits this variation, so P1 applies to all nine.
 
\paragraph{Principle 2 (P2): Environment alignment.}
Several capabilities reduce to inspecting properties of the agent's
artifacts (formula structure, layout metadata, cross-file references)
that do not survive flattening to text or rendering as images. Six capabilities in the taxonomy exhibit this pattern. We expect it in any environment with binary files, stateful tools, or structured data, which have analogous properties that serialization erases. Our second principle is that the verifier should run in the \emph{same environment} as the rollout agent, with access to its final state through the same filesystem, libraries, and tools (including domain-specific
Model Context Protocol~(MCP)~\citep{hou:mcp} tools). Unrestricted access would let the verifier modify what it grades. Its access must therefore be read-only, so that verification always sees the agent's original, unaltered output (\Cref{sec:gandalf}). Running in the agent's environment also makes the verifier portable, because applying it to a new environment automatically provides access to that environment's existing tools, and there is no dependency on domain-specific parsers.
 
\paragraph{Principle 3 (P3): Domain guidance.}
Several capabilities require knowledge that does not appear in the
criterion text, such as accounting sign conventions or disclosure
standards for an investment memo. Two capabilities in the taxonomy exhibit
this pattern. Every domain has analogous
conventions, such as clinical standards in medical workflows, citation rules in legal
workflows, and communication norms in productivity workflows. Our third principle is that the verifier should accept domain guidance
at deployment time through a dedicated \emph{guidance channel},
separate from the criterion text and from the architecture itself. Guidance is written once for a corpus of tasks and states reusable conventions and inspection policies.

\section{Gandalf: a verifier for rubric-graded environments}
\label{sec:gandalf}

We implement the design principles of \Cref{sec:design-principles} in \emph{Gandalf}, a portable Python application that grades a rollout with an agent, the \emph{judge}. For each criterion, the judge returns a met/unmet verdict with its reasoning and the evidence it checked (\Cref{app:gandalf}).

\paragraph{Reactive verification (P1).}
\gandalf instantiates P1 by making the judge a reactive agent, built on the open-source OpenHands~SDK~\citep{wang:openhands}. Equipped with a terminal and a file editor, the judge decides for itself which files to open and which checks to run for each criterion.

\paragraph{Environment alignment (P2).}
To satisfy P2, Gandalf is designed to run inside the rollout's environment. There, the judge opens the agent's files in their native formats, runs code with the installed libraries, and calls the environment's MCP tools.

Running in the agent's environment requires keeping the judge and the graded agent apart. The judge starts only after the rollout has finished and works on a temporary copy of the workspace under a strictly scoped separate Unix user. Its tools also cannot modify the agent's work or the environment's data, so the judge cannot change what it grades.

\paragraph{Domain guidance (P3).}
\gandalf supports P3 through a guidance channel. Domain guidance, written once for an environment rather than per task, enters the judge's prompt alongside the task instruction, the rollout agent's final message, and the criteria. 

\paragraph{Beyond the principles.}
Two further settings affect cost and accuracy. The choice of the judge's model lets a benchmark designer trade grading accuracy against cost (\Cref{sec:eval-sensitivity}). Batching lets one session grade many criteria, so the judge reuses the files it has already inspected instead of opening them again for each criterion. A rubric can also be split into several batches that are graded in parallel.

\section{Evaluation}
\label{sec:evaluation}

We evaluate Gandalf along with three baselines described in
\Cref{sec:background}: Archipelago (serialize-then-grade),
Agent-as-a-Judge (AAAJ; workflow-based file inspection), and AutoRubric
(trajectory-only grading without access to the deliverable). The evaluation answers three questions:
(1) how do verifiers compare on accuracy and cost
(\Cref{sec:verifier-comparison});
(2) what do Gandalf's domain guidance and MCP tools contribute
(\Cref{sec:eval-principles}); and
(3) how do batch size and backing model trade accuracy against cost
(\Cref{sec:eval-sensitivity}).

\paragraph{Setup.}
In the verifier and model comparisons, Gandalf grades each task's rubric in two agent sessions, each covering approximately half the criteria. \Cref{sec:eval-sensitivity} examines how batch size and backing model affect F1 and model API cost. Gandalf receives domain guidance and access to the environment-aligned MCP tools, using Gemini~3~Flash unless stated otherwise.

\paragraph{Metrics.}
In \benchmarkabbrev, 66.3\% of criterion–rollout pairs are met, so a verifier that always predicts ``met'' reaches 66.3\% accuracy without detecting any unmet criteria. We therefore report F1 on the unmet class (UNMET F1) as our main evaluation metric; throughout the paper, F1 refers to UNMET F1. It is zero for the always-``met'' verifier and rewards both finding unmet criteria (recall) and avoiding false alarms on met ones (precision). We evaluate every task three times per configuration, or once where indicated, and compute a single F1 over all criterion verdicts from these evaluations. Cost is the average model API cost per complete dataset pass over all 24 tasks, reflecting provider prompt-caching discounts and excluding sandbox compute, document extraction, and failed attempts.

\subsection{Gandalf leads the cost–F1 frontier}
\label{sec:verifier-comparison}
\begin{figure*}[t]
\centering
\begingroup
\colorlet{archipelagoMain}{archipelagoMain!72!black}
\colorlet{aajMain}{aajMain!82!black}
\pgfplotstableread[col sep=comma]{shared/figures/data/cost_f1.csv}\costdata
\pgfkeys{/costlabel/.cd,
  M01/.style={anchor=south east,xshift=-3pt,yshift=3pt},
  M02/.style={anchor=south east,xshift=-12pt,yshift=9pt},
  M03/.style={anchor=north west,xshift=3pt,yshift=-4pt},
  M04/.style={anchor=south west,xshift=9pt,yshift=12pt},
  M05/.style={anchor=south west,xshift=3pt,yshift=3pt},
  M06/.style={anchor=south west,xshift=3pt,yshift=3pt},
  M07/.style={anchor=south,xshift=0pt,yshift=12pt},
  M08/.style={anchor=north west,xshift=3pt,yshift=-4pt},
  M13/.style={anchor=north east,xshift=-10pt,yshift=-16pt},
  M14/.style={anchor=north east,xshift=-3pt,yshift=-4pt},
  M15/.style={anchor=south east,xshift=-3pt,yshift=3pt},
  M16/.style={anchor=north west,xshift=3pt,yshift=-4pt},
  M17/.style={anchor=south west,xshift=3pt,yshift=3pt},
  M18/.style={anchor=south west,xshift=3pt,yshift=3pt},
  M19/.style={anchor=west,xshift=8pt,yshift=0pt},
}
\pgfkeys{/costconnector/.cd,
  M02/.initial=south east,
  M04/.initial=south west,
  M13/.initial=north east,
}
\def\costvalue#1#2{\pgfplotstablegetelem{#1}{#2}\of\costdata}
\newcommand{\costseries}[5]{%
  \def\selectedverifier{#1}\def\selectedrepeats{#4}%
  \addplot[
    only marks,solid,color=#2,mark=#3,mark size=2.0pt,
    mark options={solid,fill=#5,draw=#2},forget plot,
    x filter/.code={%
      \pgfplotstablegetelem{\coordindex}{Family}\of\costdata
      \edef\rowfamily{\pgfplotsretval}%
      \pgfplotstablegetelem{\coordindex}{Verifier}\of\costdata
      \edef\rowverifier{\pgfplotsretval}%
      \pgfplotstablegetelem{\coordindex}{Repeats}\of\costdata
      \edef\rowrepeats{\pgfplotsretval}%
      \ifx\rowfamily\panelfamily\else\def\pgfmathresult{nan}\fi
      \ifx\rowverifier\selectedverifier\else\def\pgfmathresult{nan}\fi
      \ifx\rowrepeats\selectedrepeats\else\def\pgfmathresult{nan}\fi
    },
    error bars/y dir=both,error bars/y explicit,
    error bars/error bar style={solid,line width=0.45pt,draw opacity=0.85},
    error bars/error mark options={rotate=90,mark size=1.5pt},
    mark layer=axis foreground,
  ] table[x=Cost,y=F1,y error plus=ErrPlus,y error minus=ErrMinus]{\costdata};
}
\newcommand{\geminicostseries}{%
  \costseries{gandalf}{gandalfMain}{*}{3}{gandalfMain}
  \costseries{gandalf}{gandalfMain}{*}{1}{white}
  \costseries{archipelago}{archipelagoMain}{triangle*}{3}{archipelagoMain}
  \costseries{aaaj}{aajMain}{square*}{3}{aajMain}
  \costseries{autorubric}{autorubricMain}{diamond*}{3}{autorubricMain}
}
\newcommand{\gptcostseries}{%
  \costseries{gandalf}{gandalfMain}{*}{3}{gandalfMain}
  \costseries{archipelago}{archipelagoMain}{triangle*}{3}{archipelagoMain}
  \costseries{aaaj}{aajMain}{square*}{1}{white}
}
\def\costlabelrow#1{%
  \costvalue{#1}{Family}\edef\rowfamily{\pgfplotsretval}%
  \ifx\rowfamily\panelfamily
    \costvalue{#1}{Verifier}\edef\verifier{\pgfplotsretval}%
    \def\target{gandalf}\ifx\verifier\target\def\pointcolor{gandalfMain}\fi
    \def\target{archipelago}\ifx\verifier\target\def\pointcolor{archipelagoMain}\fi
    \def\target{aaaj}\ifx\verifier\target\def\pointcolor{aajMain}\fi
    \def\target{autorubric}\ifx\verifier\target\def\pointcolor{autorubricMain}\fi
    \costvalue{#1}{Cost}\edef\pointx{\pgfplotsretval}%
    \costvalue{#1}{F1}\edef\pointy{\pgfplotsretval}%
    \costvalue{#1}{Label}\edef\pointlabel{\pgfplotsretval}%
    \costvalue{#1}{Condition}\edef\condition{\pgfplotsretval}%
    \edef\drawlabel{\noexpand\node[font=\noexpand\scriptsize,text=\pointcolor,inner sep=1pt,
      /costlabel/\condition] (costlabel\condition) at (axis cs:\pointx,\pointy) {\pointlabel};}
    \drawlabel
    \pgfkeysifdefined{/costconnector/\condition}{%
      \pgfkeysgetvalue{/costconnector/\condition}\connectoranchor
      \edef\drawconnector{\noexpand\draw[draw=\pointcolor!75,line width=0.35pt,
        shorten <=3pt,shorten >=1pt]
        (axis cs:\pointx,\pointy) -- (costlabel\condition.\connectoranchor);}
      \drawconnector
    }{}
  \fi
}
\pgfplotstablegetrowsof{\costdata}\pgfmathtruncatemacro{\lastrow}{\pgfplotsretval-1}
\pgfplotstablesort[sort key=Cost]\sortedcostdata{\costdata}
\newcommand{\costfrontier}{%
  \def\frontiercoords{}\def\bestfscore{-1}%
  \pgfplotsinvokeforeach{0,...,\lastrow}{%
    \pgfplotstablegetelem{##1}{Family}\of\sortedcostdata
    \edef\rowfamily{\pgfplotsretval}%
    \ifx\rowfamily\panelfamily
      \pgfplotstablegetelem{##1}{F1}\of\sortedcostdata
      \edef\frontierfscore{\pgfplotsretval}%
      \pgfmathparse{\frontierfscore>\bestfscore}%
      \ifnum\pgfmathresult=1\relax
        \pgfplotstablegetelem{##1}{Cost}\of\sortedcostdata
        \edef\frontiercoords{\frontiercoords (\pgfplotsretval,\frontierfscore)}%
        \let\bestfscore\frontierfscore
      \fi
    \fi
  }%
  \edef\drawfrontier{\noexpand\addplot[
    no marks,dashed,color=black!55,line width=0.8pt,forget plot]
    coordinates {\frontiercoords};}%
  \drawfrontier
}
\begin{tikzpicture}
\pgfplotsset{set layers=standard}
\begin{groupplot}[
  group style={group size=2 by 1,horizontal sep=1.0cm},
  width=0.49\linewidth,height=0.43\linewidth,
  xmode=log,log basis x=10,log ticks with fixed point,
  xmin=4,xmax=220,xtick={5,10,20,50,100,200},
  ymin=0.50,ymax=0.97,ytick={0.5,0.6,0.7,0.8,0.9},
  xlabel={Model API cost per dataset pass\\(USD, log scale)},
  xlabel style={align=center},
  tick label style={font=\scriptsize},label style={font=\small},
  title style={font=\small\bfseries},
  ymajorgrids=true,grid style={gray!18},axis line style={black!50},
  tick align=outside,tick pos=left,clip=false,
]
\nextgroupplot[title={Gemini models},ylabel={UNMET F1}]
\def\panelfamily{Gemini}
\costfrontier
\geminicostseries
\pgfplotsinvokeforeach{0,...,\lastrow}{\costlabelrow{#1}}
\nextgroupplot[title={GPT models},yticklabels={}]
\def\panelfamily{GPT}
\costfrontier
\gptcostseries
\pgfplotsinvokeforeach{0,...,\lastrow}{\costlabelrow{#1}}
\end{groupplot}
\matrix[column sep=22pt,anchor=north,font=\scriptsize,
  nodes={inner sep=0pt}] (costlegend)
 at ($(group c1r1.south)!0.5!(group c2r1.south)+(0,-1.5cm)$) {
  \node(legendgan){Gandalf}; & \node(legendarc){Archipelago}; &
  \node(legendaaaj){Agent-as-a-Judge}; & \node(legendauto){AutoRubric}; \\
};
\foreach \legendnode/\legendcolor/\legendmark in {
  legendgan/gandalfMain/*,legendarc/archipelagoMain/triangle*,
  legendaaaj/aajMain/square*,legendauto/autorubricMain/diamond*}{%
  \path ([xshift=-7pt]\legendnode.west)
    node[inner sep=0pt,text height=0pt,text depth=0pt,
      text=\legendcolor,anchor=center]
      {\pgfsetplotmarksize{2pt}\pgfuseplotmark{\legendmark}};
}
\node[anchor=north,font=\scriptsize,inner sep=0pt]
  at ([yshift=-4pt]costlegend.south) {%
  \tikz[baseline=-0.5ex]\draw[black!75,fill=black!75] (0,0) circle (1.8pt);%
  \enspace Filled: 3 evaluations/task\qquad
  \tikz[baseline=-0.5ex]\draw[black!75,fill=white] (0,0) circle (1.8pt);%
  \enspace Hollow: 1 evaluation/task};
\end{tikzpicture}
\endgroup
\caption{UNMET F1 and model API cost per pass on \benchmarkabbrev\ (24 tasks), grouped by backing-model family. Dashed lines connect the Pareto-optimal configurations in each panel based on point estimates. Error bars show 95\% confidence intervals from resampling tasks; they do not separately estimate variation between repeated evaluations. AutoRubric with GPT-5.4 was stopped early because of cost. For Gandalf, per-pass cost reflects token use, not per-token price alone: Gemini~3.8~Flash made about $1.5\times$ as many LLM calls as Gemini~3.1~Pro, with larger contexts and more reasoning per call.}
\label{fig:model-vary-split}
\end{figure*}
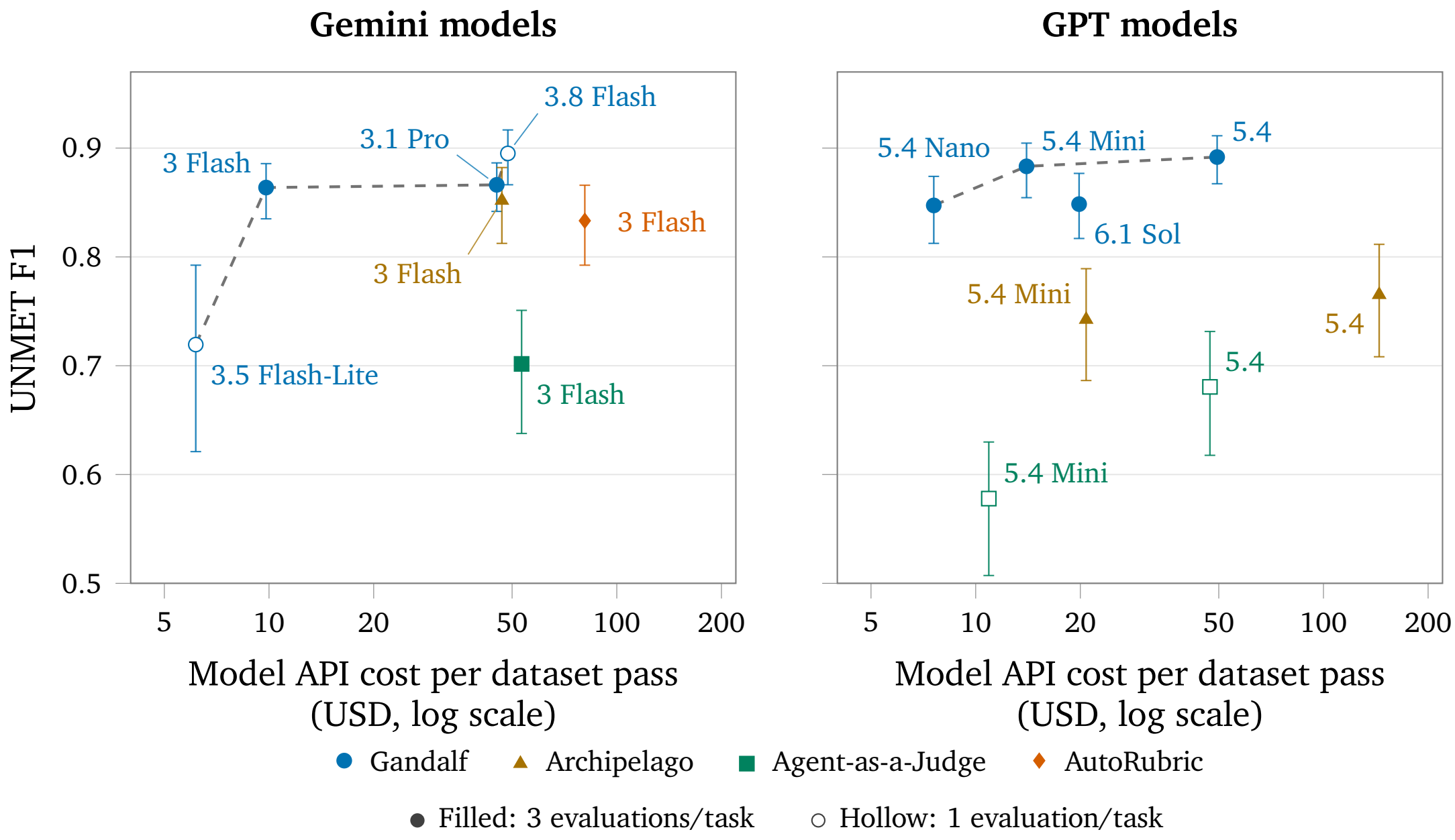

Across two model families, Gemini and GPT, Gandalf offers configurations that match or exceed each baseline's observed F1 at lower model API cost (\Cref{fig:model-vary-split}). With Gemini~3~Flash, Gandalf achieves F1~0.864 at \$9.81 per dataset pass, compared with Archipelago's 0.852 at \$46.73: similar F1 at about one-fifth the cost. With Gemini~3~Flash, Gandalf also achieves higher F1 at lower cost than AAAJ and AutoRubric.\footnote{AutoRubric with GPT-5.4 was stopped because of cost and is excluded from the full-dataset comparison. A single task cost \$189: resending the full trajectory for every criterion required about 357K input tokens per request.}

With GPT models, Gandalf improves on Archipelago in both F1 and cost. With GPT-5.4-Mini, Gandalf leads by 0.141 F1 (0.883 vs.\ 0.742) at two-thirds the cost (\$14.00 vs.\ \$20.78). With GPT-5.4, it leads by 0.127 F1 (0.892 vs.\ 0.765) at one-third the cost (\$49.46 vs.\ \$144.64). Both paired 95\% intervals exclude zero. Even Gandalf's cheapest GPT configuration (GPT-5.4-Nano, F1~0.847, \$7.58) exceeds the F1 of Archipelago with GPT-5.4 at one-nineteenth the cost.

\subsection{Design-principle ablations}
\label{sec:eval-principles}
\newcommand{\gandalfdesignrows}{%
\cmark & \cmark & .877 & \$25.11 \\
\cmark & \xmark & .881 & \$24.93 \\
\xmark & \cmark & .873 & \$22.17 \\
\xmark & \xmark & .877 & \$24.81 \\
}
\newcommand{\gandalfbatchrows}{%
64 & .858 & \$9.87 \\
$\star$ $\approx$51 (2 sessions) & .862 & \$10.16 \\
32 & .873 & \$15.52 \\
16 & .880 & \$25.72 \\
8 & .874 & \$43.10 \\
4 & .870 & \$71.85 \\
}
\newcommand{\gandalfmodelrows}{%
Gemini 3.8 Flash$^\dagger$ & .895 & \$48.62 \\
Gemini 3.1 Pro & .866 & \$45.17 \\
$\star$ Gemini 3 Flash & .864 & \$9.81 \\
Gemini 3.5 Flash-Lite$^\dagger$ & .719 & \$6.16 \\
GPT-5.4 & .892 & \$49.46 \\
GPT-6.1-Sol & .849 & \$19.83 \\
GPT-5.4-Mini & .883 & \$14.00 \\
GPT-5.4-Nano & .847 & \$7.58 \\
}

\newcommand{\tbsep}{\hskip 3.5pt\vrule\hskip 3.5pt}
\newsavebox{\gandalfdesignbox}
\newsavebox{\gandalfbatchbox}
\newsavebox{\gandalfmodelbox}

\begin{table*}[t]
\caption{Gandalf ablations on \benchmarkabbrev. (a)~Design principles:
guidance (P3) and MCP tools (one part of P2) switched on and off, in a separate study with 16
criteria per session; compare rows only within~(a). (b)~Batch size
(criteria per session) and (c)~model, each varied from the default ($\star$:
Gemini~3~Flash, guidance, MCP, two agent sessions per task). F1 treats UNMET as the
positive class; cost is model API cost per dataset pass;
$^\dagger$one pass, otherwise three.}
\label{tab:gandalf-ablations}
\centering
\small
\setlength{\tabcolsep}{3pt}
\renewcommand{\arraystretch}{1.2}
\captionsetup[subtable]{position=top,skip=2pt}
\sbox{\gandalfdesignbox}{%
\begin{tabular}[t]{@{} c c c r @{}}
\toprule
\begin{tabular}[b]{@{}c@{}}MCP\\tools\end{tabular} &
\begin{tabular}[b]{@{}c@{}}Guidance\\(P3)\end{tabular} & F1 & Cost \\
\midrule
\gandalfdesignrows
\bottomrule
\end{tabular}%
}
\sbox{\gandalfbatchbox}{%
\begin{tabular}[t]{@{} l c r @{}}
\toprule
\begin{tabular}[b]{@{}l@{}}Batch\\size\end{tabular} & F1 & Cost \\
\midrule
\gandalfbatchrows
\bottomrule
\end{tabular}%
}
\sbox{\gandalfmodelbox}{%
\begin{tabular}[t]{@{} l c r @{}}
\toprule
Model & F1 & Cost \\
\midrule
\gandalfmodelrows
\bottomrule
\end{tabular}%
}
\begin{subtable}[t]{\wd\gandalfdesignbox}
\caption{Design principles}\label{tab:abl-design}
\usebox{\gandalfdesignbox}
\end{subtable}\hfill
\begin{subtable}[t]{\wd\gandalfbatchbox}
\caption{Batch size}\label{tab:abl-batch}
\usebox{\gandalfbatchbox}
\end{subtable}\hfill
\begin{subtable}[t]{\wd\gandalfmodelbox}
\caption{Model}\label{tab:abl-model}
\usebox{\gandalfmodelbox}
\end{subtable}
\end{table*}

In \Cref{tab:abl-design}, we examine the contribution of P2
(environment alignment) and P3 (judge guidance) by toggling MCP tools
and guidance in turn, fixing the base model to Gemini~3~Flash and
batch size to 16 in a separate study. These ablations measure the
incremental effects of MCP and guidance within a P1-equipped verifier
that retains terminal access to the workspace, rather than the full
impact of environment access or reactive verification.
 
No F1 effect is resolved: the cells span 0.008~F1, and the 95\% CIs of
both main effects and their interaction include zero. Guidance instead
makes Gandalf stricter: averaged across MCP on and off, it predicts UNMET
2.2~percentage points more often ($[1.2,3.2]$), lowering precision by
0.030 ($[-0.045,-0.014]$) and raising recall by 0.024 ($[0.009,0.043]$), a shift
\Cref{app:rewardkit} also finds with GPT-5.4-Mini. MCP adds \$2.94 per
pass with guidance ($[1.99,4.09]$) and \$0.12 without ($[-0.55,0.81]$).
One likely reason for the null is redundancy: the agent's own MCP calls
already copy the market and filing files they retrieve into the
workspace the judge inspects, and many criteria state the expected value
(e.g., ``within $\pm$5\% of \$102.1bn''). The study thus resolves no F1
benefit from MCP tools, one part of P2, or from this guidance; we expect
both to matter where criteria leave source data or domain conventions
implicit.

\subsection{Sensitivity to batch size and model choice}
\label{sec:eval-sensitivity}

\paragraph{Batch size.}
In \Cref{tab:abl-batch}, we vary batch size, the number of criteria graded by one agent
trajectory. F1 rises from 0.858 at $b{=}64$ to 0.880 at $b{=}16$,
then falls to 0.870 at $b{=}4$, with cost rising from \$9.87 to
\$71.85 across the full range. Below $b{=}16$, additional sessions
increase cost without improving observed F1: dropping from $b{=}16$
to $b{=}4$ costs 2.8$\times$ as much (\$25.72 vs.\ \$71.85).
We use two sessions per task for all cross-verifier comparisons as a
deliberate cost-accuracy tradeoff; in the batch-size study, this
configuration achieves F1~0.862 at \$10.16 per pass.
 
\paragraph{Model choice.}
\Cref{tab:abl-model} also reports Gandalf with two sessions
per task across eight models spanning two providers. Gemini~3.8~Flash
achieves the highest observed F1 (0.895, one pass), followed by
GPT-5.4 (0.892) and GPT-5.4-Mini (0.883). The full spread from the
weakest model (Gemini~3.5~Flash-Lite, 0.719, one pass) to the strongest
is 0.176 F1; among models evaluated over three passes, the
spread is 0.045 F1 (0.847--0.892). Model choice therefore matters,
although several Gandalf configurations offer a strong cost--F1
tradeoff (\Cref{sec:verifier-comparison}).

\section{Discussion}
\label{sec:discussion}

We now step back and ask two questions about the evaluation results: (1) does \gandalf generalize beyond
\benchmarkabbrev (\Cref{sec:discussion-generalize}); and (2) why do
different verifiers perform differently
(\Cref{sec:discussion-capability})?

\subsection{\gandalf generalizes beyond \benchmarkabbrev}
\label{sec:discussion-generalize}

\newcommand{\openclawrows}{%
Gandalf & Gemini 3 Flash & \textbf{0.955} & \$7.68 \\
\midrule
AutoRubric & Gemini 3 Flash & 0.934 & \$1.43 \\
Archipelago & Gemini 3 Flash & 0.842 & \$1.53 \\
Archipelago & GPT-5.4-Mini & 0.841 & \$0.70 \\
AAAJ & Gemini 3 Flash & 0.827 & \$12.12 \\
AAAJ & GPT-5.4-Mini & 0.747 & \$1.11 \\
}
\newcommand{\openclawunmetshare}{59.6\%}
\newcommand{\openclawalwaysunmetfone}{0.747}
\newcommand{\bvbunmetshare}{33.7\%}
\newcommand{\bvbalwaysunmetfone}{0.504}

To investigate whether \gandalf generalizes, we collected another internal dataset analogous to \benchmarkabbrev in a new domain: we additionally evaluate on an OpenClaw~(OPC) benchmark of persona-driven
personal-productivity workflows: managing schedules, coordinating
with contacts, planning lifestyle tasks. OpenClaw differs from \benchmarkabbrev
in domain (consumer), tool surface
(messaging clients, calendar and booking services, a weather API), and evidence type (many OpenClaw tools are stateful, so verification depends on determining if the environment has changed such as whether a message was sent or an event created). This dataset consists of 396 criteria judgments across 24 tasks. As on \benchmarkabbrev, we report UNMET F1. Most OpenClaw criteria are unmet (\openclawunmetshare, vs.\ \bvbunmetshare\ on \benchmarkabbrev), so a verifier that always predicts unmet scores \openclawalwaysunmetfone\ here and \bvbalwaysunmetfone\ on \benchmarkabbrev; absolute F1 values are therefore not directly comparable across the two datasets. \Cref{app:error-direction} reports confusion counts for the Gemini~3~Flash verifiers.

\skipkdd{%
\begin{table}[b]
\caption{Verifier F1 and model API cost per dataset pass on OpenClaw
(24~episodes, three passes).}
\label{tab:openclaw-results}
\centering
\small
\setlength{\tabcolsep}{6pt}
\renewcommand{\arraystretch}{1.2}
\begin{tabular}{llcr}
\toprule
Verifier & Model & F1 & Cost \\
\midrule
\openclawrows
\bottomrule
\end{tabular}
\end{table}
}
\onlykdd{%
\begin{figure*}[t]
\centering
\includegraphics[width=0.76\textwidth]{shared/figures/blog-post/plot_openclaw.pdf}
\caption{Cost vs.\ F1 across verifiers on OpenClaw (24~episodes), with Gemini~3~Flash (circles) and GPT-5.4~Mini (triangles) model families. Gandalf with Gemini~3~Flash sits at the Pareto frontier (F1~0.951, \$8.20).}
\label{fig:openclaw-results}
\end{figure*}
}

\skipkdd{\Cref{tab:openclaw-results}}\onlykdd{\Cref{fig:openclaw-results}}
reports F1 and cost across all four
verifiers on the OpenClaw dataset.
Gandalf achieves F1 of 0.955, leading the next-best verifier
(AutoRubric with Gemini~3~Flash, 0.934) by 0.021 F1, and is
again Pareto optimal on observed F1 and model API cost. AutoRubric
costs \$1.43 per pass against Gandalf's \$7.68; Archipelago with
GPT-5.4-Mini is cheaper still (\$0.70, F1~0.841). Gandalf with
Gemini~3~Flash has the highest observed F1 among the evaluated
configurations, but AutoRubric's competitive result shows that the
accuracy--cost tradeoff depends on the environment.

\subsection{Verifiers' errors are consistent with gaps in capabilities}
\label{sec:discussion-capability}
\Cref{tab:combined} breaks down errors by capability taxonomy generated in \Cref{sec:taxonomy}. The error patterns are consistent with gaps in the capabilities each verifier supports, while Gandalf has the lowest observed error on seven of the nine BVB capabilities and five of the six measured on OpenClaw.

\newcommand{\combinedrows}{%
Audit quantitative metric calculation & 522 & \textbf{7.0} & 9.0 & 19.3 & 11.8 & - & - & - & - & - \\
Inspect visual presentation and style & 406 & \textbf{16.1} & 16.2 & 25.0 & 17.0 & - & - & - & - & - \\
Verify literal data point accuracy & 399 & \textbf{4.8} & 5.3 & 11.8 & 5.9 & 200 & \textbf{3.0} & 14.7 & 20.8 & 4.3 \\
Verify structural component inventory & 322 & \textbf{4.9} & 7.0 & 13.6 & 7.0 & 53 & \textbf{1.3} & 18.2 & 7.5 & 5.7 \\
Reconcile cross-component data linkage & 227 & 15.4 & 14.5 & 28.8 & \textbf{14.1} & 20 & \textbf{10.0} & 36.7 & 30.0 & 20.0 \\
Validate document labeling and metadata & 223 & \textbf{9.0} & 10.3 & 16.0 & 9.7 & 41 & 6.5 & 31.7 & 11.4 & \textbf{2.4} \\
Audit governance and strategic narrative & 129 & 10.3 & \textbf{8.0} & 14.5 & 9.3 & 66 & \textbf{13.6} & 34.8 & 30.3 & 17.2 \\
Validate temporal and sequential integrity & 123 & \textbf{7.0} & 14.1 & 12.2 & 11.1 & 12 & \textbf{5.6} & 33.3 & 16.7 & 16.7 \\
Audit accounting logic and convention & 113 & \textbf{10.9} & 13.0 & 23.9 & 16.2 & - & - & - & - & - \\
\midrule
Aggregate & 2{,}464 & \textbf{9.2} & 10.3 & 18.5 & 11.1 & 396 & \textbf{5.3} & 21.8 & 19.8 & 7.6 \\
}

\begin{table*}[t]
\caption{Per-capability error rates (\%) for Gandalf (Gan), Archipelago (Arc), Agent-as-a-Judge (AAAJ), and AutoRubric (Aut) on \benchmarkabbrev and OpenClaw. Bold: best verifier on each row within
each corpus (ties bolded). Cells marked {-} denote capabilities with
fewer than five OPC criteria.}
\label{tab:combined}
\centering
\small
\setlength{\tabcolsep}{4pt}
\renewcommand{\arraystretch}{1.15}
\begin{tabular}{>{\raggedright\arraybackslash}p{3.6cm}rrrrr@{\hskip 6pt}rrrrr}
\toprule
& \multicolumn{5}{c}{\benchmarkabbrev} & \multicolumn{5}{c}{OpenClaw} \\
\cmidrule(lr){2-6}\cmidrule(lr){7-11}
Capability & $n$ & Gan. & Arc. & AAAJ & Aut. & $n$ & Gan. & Arc. & AAAJ & Aut. \\
\midrule
\combinedrows
\bottomrule
\end{tabular}
\end{table*}

\paragraph{AutoRubric.}
AutoRubric has no artifact access and no domain guidance. It grades
each criterion in an isolated LLM call over the agent's trajectory.
Its errors concentrate on capabilities where the criterion depends on how the deliverable turned out, not on what the agent did to produce it. 
On \benchmarkabbrev{}, errors are
lowest on \emph{verify literal data point accuracy} (5.9\%) and
\emph{verify structural component inventory} (7.0\%), capabilities
where a specific tool call maps to a specific check: writing a value
or creating a section is recorded in the trajectory. Errors are
highest on \emph{inspect visual presentation and style} (17.0\%) and
\emph{audit accounting logic and convention} (16.2\%), capabilities
where the criterion depends on properties of the final rendered or
structured output that no single tool call captures. The same split
holds on OPC, where errors are also high on \emph{reconcile
cross-component data linkage} (20.0\%) and \emph{audit governance
and strategic narrative} (17.2\%).
 
\paragraph{AAAJ.}
AAAJ inspects files through its own toolset rather than the rollout's
environment, with no domain guidance. It is a workflow-based (rather than reactive)
agent, lacking all of P1--P3.
Its errors concentrate on capabilities where the
deliverable's structured internals carry the signal. On
\benchmarkabbrev{}, \emph{reconcile cross-component data linkage}
(28.8\%) and \emph{inspect visual presentation and style} (25.0\%)
are the highest-error capabilities. Both require access to
structured internals, formula references for the former and layout
metadata for the latter, which general-purpose file reading
extracts only with substantial effort. On
OPC, AAAJ reaches 30.0\% on \emph{reconcile
cross-component data linkage} and 30.3\% on \emph{audit governance
and strategic narrative}, roughly two to three times Gandalf's error
on the same capabilities.
 
\paragraph{Archipelago.}
Archipelago is workflow-based and grades each criterion from a
serialized snapshot of the deliverable, giving it a partial form of
P2. It has neither P1 nor P3. On \benchmarkabbrev, its aggregate
error (10.3\%) is close to Gandalf's (9.2\%), consistent with
serialization capturing much of the evidence needed on this corpus.
The gap widens sharply on OPC, where Archipelago's aggregate error
jumps to 21.8\% against Gandalf's 5.3\%. The gap is large on
capabilities involving OPC's stateful tools, consistent with
limitations of serialized evidence: \emph{reconcile cross-component data linkage}
(36.7\% vs 10.0\%), \emph{audit governance and strategic narrative}
(34.8\% vs 13.6\%), and \emph{validate temporal and sequential
integrity} (33.3\% vs 5.6\%). This pattern is consistent with
serialized snapshots being less effective when evidence depends on
what the agent's tools did to the environment.

\paragraph{Where Gandalf does not win.}
Gandalf does not win on two \benchmarkabbrev{} capabilities. It
loses narrowly on \emph{reconcile cross-component data linkage} (by
1.3~percentage points to AutoRubric and 0.9 to Archipelago) and
\emph{audit governance and strategic narrative} (by 2.3~percentage points to
Archipelago). It narrowly leads Archipelago on \emph{inspect visual
presentation and style} (16.1\% vs.\ 16.2\%). The small differences
on narrative and presentation are consistent with these capabilities
being evaluable from rendered output alone. The loss on
\emph{reconcile cross-component data linkage} shows a different
limit: the capability is hard for every verifier (errors range from
14.1\% to 28.8\%), suggesting that environment access alone does not
resolve the difficulty of tracing values across files.
OPC provides evidence consistent with the first pattern. On
\emph{audit governance and strategic narrative}, OPC's criteria
require inspecting what stateful tools did (for example, whether a
message was sent), not judging rendered text. Here Gandalf's error
is 21.2~percentage points lower than Archipelago's (13.6\% vs.\
34.8\%), consistent with the value of environment access when
criteria require inspecting tool-mediated state.

\section{Related work}

\paragraph{Agent environments.} Agent benchmarks frequently rely on deterministic checks against final state, expected tool calls, or unit tests~\citep{wang:gta,li:toolathlon,patil:bfcl,qin:toolllm,chen:t-eval,li:api-bank,patil:gorilla,yao:tau-bench,liu:agentbench,jimenez:swe-bench,li:clawsbench,malay:enterpriseops-gym}. In pursuit of greater scope and realism, recent benchmarks in semi-verifiable domains have embraced rubric-based grading. Such benchmarks include BankerToolBench~\citepbtb, APEX-Agents~\citep{vidgen:apex-agents}, CoreCraft~\citep{mehta:corecraft}, \mbox{GTA-2}~\citep{wang:gta-2}, ClawBench~\citep{zhang:clawbench}, Claw-Eval~\citep{ye:claw-eval}, and AgencyBench~\citep{li:agencybench}. Across both lines, these works treat the verifier as a means to an end, not the object of study.

\paragraph{Rubrics.} Rubrics are a popular approach to defining correctness in semi-verifiable domains~\citep{lin:wildbench,deshpande:multichallenge}. Rubrics may be authored by domain experts~\citep{arora:healthbench,starace:paperbench,akyurek:prbench} or generated synthetically~\citep{cook:tick,liu:openrubrics,wadhwa:evalagent,raghavendra:agentic-rubrics}. RubricBench~\citep{zhang:rubricbench} evaluates rubric generators themselves against human preferences. A line of work investigates rubrics in the context of training~\citep{viswanathan:rlcf,gunjal:rar}; some of this work investigates training with synthetic rubrics~\citep{rezaei:onlinerubrics,shao:drtulu,huang:rubicon,he:rifl}. Rubric-ARM~\citep{xu:rubric-arm} jointly optimizes a rubric generator and a judge, optimizing judgment accuracy for matching human preference ranking. Unlike these works, we take rubrics as a given and study the verifiers that consume them.

\paragraph{Verifier designs.} Verifiers, both inside and outside the context of environments, cover a broad design space~\citep{guan:search-verify-feedback,li:llm-as-a-judge-survey,you:agent-as-a-judge-survey}. A line of work optimizes LLM-as-a-judge for evaluating text through prompting or training~\citep{liu:g-eval,kim:prometheus,kim:prometheus-2,zhu:judgelm}, and a related line decomposes evaluation into per-criterion judgments via rubrics~\citep{hashemi:llm-rubric,lee:checkeval,rao:autorubric,zhou:autochecklist,saad-falcon:lmunit}. Benchmarks commonly apply such techniques for outcome grading in agent environments by evaluating a serialized trajectory or serialized output files~\citep{mehta:corecraft,ye:claw-eval}. In contrast, some environments propose workflow-based agents as verifiers~\citep{zhuge:agent-as-a-judge,vidgen:apex-agents,zhu:findeepresearch,findeis:agent-evaluator,peng:rewardagent,han:verifiagent,sun:finresearchbench}, while others propose reactive agents~\citep{lin:jade,avraham:dream,cui:vagen,dai:prore,wang:codevisionary,ding:arm-thinker,xu:tir-judge,ma:agentic-verifier}. Despite the broad design space, no principled account describes which design choices matter for which verification capabilities. Concurrently with this work, RewardKit in Harbor~\citep{harbor} adopted a similar reactive, in-environment verifier design; \Cref{app:rewardkit} compares the two at matched models.

\paragraph{Meta-evaluation of verifiers.} Motivated by the importance of verifiers in benchmarks, meta-evaluation benchmarks evaluate verifiers themselves, covering LLM-as-a-judge for preference rankings~\citep{zheng:arena,tan:judgebench,lambert:rewardbench,malik:rewardbench-2} and LLM-as-a-judge for rubric grading~\citep{pan:rubriceval}. Verifiers can be domain-specific, motivating domain-specific benchmarks such as AgentRewardBench~\citep{lu:agentrewardbench} for web agents and meta-evaluations such as TinyV~\citep{xu:tinyv} and \citet{huang:rl-verifier-robustness} for mathematics. \benchmarkabbrev is the first meta-evaluation dataset that covers rubric-graded agent environments, where rubric criteria cover deliverable files and side-effecting tool calls.

\section{Conclusion}
\label{sec:conclusion}

Gandalf offers a strong accuracy--cost tradeoff across two rubric-graded environments. Our results motivate reactive verification, while leaving the conditions under which guidance and environment-aligned tools improve performance open for further study. We hope that the public release of \benchmarkabbrev will help facilitate continued research on building verifiers for agents, and that the open-source release of \gandalf will unlock more realistic benchmarks for agents.

\paragraph{Limitations.}
The \benchmarkabbrev dataset contains a single rollout per task; we did not evaluate the sensitivity of verifiers to the rollout distribution.
Our study measures verifier F1 score against human labels; we did not evaluate whether verifier accuracy advantage translates to better-trained agents through RL.
Our study covers 24 tasks in each of two environments; the internal OpenClaw dataset is not publicly reproducible. Cross-verifier comparisons evaluate whole systems and do not isolate the causal contribution of individual design principles. Generalization beyond these settings remains to be established.

\section*{Acknowledgments}
We thank the investment bankers who annotated and adjudicated the \benchmarkabbrev{} judgments.
This work received no third-party funding. The authors declare no competing interests beyond the affiliations listed above.

\clearpage

\printbibliography

\clearpage
\appendix
\crefalias{section}{appendix}
\crefalias{subsection}{appendix}
\crefalias{subsubsection}{appendix}
\setcounter{figure}{0}
\setcounter{table}{0}
\renewcommand{\thefigure}{A\arabic{figure}}
\renewcommand{\thetable}{A\arabic{table}}
\def\theHfigure{A\arabic{figure}}
\def\theHtable{A\arabic{table}}

\begin{center}
{\LARGE \textbf{Appendix} }
\end{center}

\section{Dataset details}
\label{app:dataset}

\subsection{Rollout provenance}
\label{app:rollouts}

\newcommand{\rolloutgenerationrows}{%
Gemini 2.5 Pro & 12 & 21.1 (3--44) & 22.2 & 23.1M & 73.5\% & 103K & \$11.50 & \$0.96 & \$0.50 \\
GLM-5 & 12 & 29.9 (11--52) & 26.3 & 12.5M & 88.2\% & 301K & \$2.86 & \$0.24 & \$0.17 \\
\midrule
All & 24 & 25.5 (3--52) & 24.2 & 35.6M & 78.6\% & 405K & \$14.36 & \$0.60 & \$0.29 \\
}
\newcommand{\rolloutverificationrows}{%
Gandalf & \$9.81 & 0.7$\times$ \\
Archipelago & \$46.73 & 3.3$\times$ \\
AAAJ & \$53.25 & 3.7$\times$ \\
AutoRubric & \$80.92 & 5.6$\times$ \\
}

\begin{table}[ht]
\caption{Generation usage of the 24 released rollouts (OpenCode 1.17.9 via OpenRouter). Tool calls are the mean (range) per rollout and steps the mean per rollout; token counts are totals, and cached is the share of prompt tokens served from the prompt cache. Cost is the recorded model API cost, reflecting prompt-caching discounts.}
\label{tab:rollout-generation}
\centering
\small
\setlength{\tabcolsep}{3.5pt}
\begin{tabular}{lrcrrrrrrr}
\toprule
& & & & \multicolumn{3}{c}{Tokens} & \multicolumn{3}{c}{Cost} \\
\cmidrule(lr){5-7}\cmidrule(lr){8-10}
Model & Rollouts & Tool calls & Steps & Prompt & Cached & Completion & Total & Mean & Median \\
\midrule
\rolloutgenerationrows
\bottomrule
\end{tabular}
\end{table}

\begin{table}[ht]
\caption{Model API cost of one verifier pass over the 24 tasks with Gemini~3~Flash, relative to the \$14.36 cost of generating the 24 rollouts.}
\label{tab:rollout-verification-cost}
\centering
\small
\setlength{\tabcolsep}{6pt}
\begin{tabular}{lrr}
\toprule
Verifier & Cost per pass & Relative to generation \\
\midrule
\rolloutverificationrows
\bottomrule
\end{tabular}
\end{table}

\section{Taxonomy Construction}
\label{app:taxonomy-construction}
The taxonomy was originally produced from an earlier
\benchmarkabbrev{} corpus using the three stages below. For the
released corpus, we retained the output of Stages~1--2 and reran
Stage~3. All LLM calls in this pipeline use Gemini~3~Flash, and
embeddings use Google's \texttt{text-embedding-005}.
 
\paragraph{Stage 1: Free-text discovery.}
For each rubric criterion, an LLM receives the agent-facing task
brief and the criterion text, and emits one to three short capability
labels along with a one-sentence rationale per label. Three
constraints are enforced programmatically:
(i) the rationale must begin with ``the verifier needs to \ldots''
and quote a phrase from the criterion text, which enforces
verifier-side framing;
(ii) labels must not contain tokens naming a specific runtime,
library, file format, or data source, which enforces
architecture-agnosticism; and
(iii) labels must follow a fixed lexical form.
 
\paragraph{Stage 2: Canonicalization.}
Each unique Stage~1 label is embedded and partitioned by
agglomerative clustering into 80 fine-grained micro-clusters. An LLM names each micro-cluster and
produces a definition, a diagnostic signal, a worked example, and a
set of non-examples contrasting it with its nearest siblings. A
final LLM call merges the 80 named micro-clusters into the canonical
taxonomy under three hard constraints: between 8 and 10 canonical
capabilities, no single capability covering more than $40\%$ of
corpus criteria, and every input micro-cluster mapped to exactly one
canonical capability.
 
\paragraph{Stage 3: Closed-vocabulary tagging.}
Every criterion is re-tagged against the closed canonical taxonomy.
An LLM receives the criterion, its task brief, and the full taxonomy
(canonical names, definitions, diagnostic signals, non-examples), and
emits one to three canonical capability tags per criterion ranked by
relevance. For the analyses in
\Cref{sec:discussion-capability}, we use the highest-ranked tag
per criterion to obtain a clean partition; the multi-tag distribution
is consistent with the primary-tag results.

\section{Gandalf implementation details}
\label{app:gandalf}

\paragraph{Execution and failure handling.}
Gandalf~1.0.0 gives each judge session the task instruction, assigned criteria
without weights, the final non-tool-call agent message extracted from the
trajectory, and optional guidance. It clones the configured work directory,
then launches an OpenHands agent with terminal, file-editor, \texttt{think},
\texttt{finish}, and configured MCP tools. Full-trajectory access is
adapter-provided: the BVB adapter places \texttt{trajectory.json} in the cloned
evidence view.

In our measured OpenHands~1.14.0 runtime, a session ends on \texttt{finish}, a
final response, stuck detection, 500 steps, or timeout. Success additionally
requires a parseable verdict covering every assigned criterion. Missing,
unparseable, or incomplete decisions become \texttt{met=None}; Gandalf retries
only those criteria. The BVB validator accepts only complete Boolean verdicts.

\begin{table}[ht]
\caption{Gandalf~1.0.0 configurations. Both environments use Gemini~3~Flash at
high effort and three evaluations per task; fixed-size BVB batching uses a
pinned partitioning extension.}
\label{tab:gandalf-config}
\centering
\footnotesize
\setlength{\tabcolsep}{3pt}
\renewcommand{\arraystretch}{1.1}
\begin{tabular}{>{\raggedright\arraybackslash}p{2.0cm}
                >{\raggedright\arraybackslash}p{3.7cm}
                >{\raggedright\arraybackslash}p{3.7cm}}
\toprule
Setting & BVB & OpenClaw \\
\midrule
Instruction and guidance &
Exact instruction; BVB guidance plus replay note &
Generic verifier instruction; no guidance. Original task available through
\texttt{trajectory.json}, not the dedicated field \\
Judge-visible evidence &
Reviewed workspace, full trajectory, final response &
Workspace, final response, and readable trajectory, rubric, and replay copies \\
MCP &
Fifteen read-only, fixed-snapshot tools &
Verifier-scoped application-state tools; mutation actions withheld \\
Sessions &
Two near-equal; at most two concurrent &
One session for all criteria \\
Timeout and Gandalf retries &
$\min(300\,\mathrm{s}\times n,1{,}740\,\mathrm{s})$; one &
$300\,\mathrm{s}\times n$; three \\
\bottomrule
\end{tabular}
\end{table}
\FloatBarrier

\subsection{Judge guidance}
\label{app:guidance}
The \href{https://github.com/Handshake-AI-Research/bankerverifierbench/blob/main/adapters/bvb/template/tests/judge-guidance.md}{BVB guidance}
is one corpus-level file for all 24 tasks; it contains inspection policies and
banking conventions, not task-specific answers. Archipelago,
Agent-as-a-Judge, and AutoRubric use native prompts without it; RewardKit
receives a shortened version (\Cref{app:rewardkit}). The factorial study uses an MCP-neutral
variant, while OpenClaw Gandalf uses no guidance.

\section{Comparison with RewardKit}
\label{app:rewardkit}

RewardKit~\citep{harbor}, the verification component of Harbor, was developed
concurrently with Gandalf. Both follow a similar verification approach: a
reactive agent judges criteria inside the task environment and inspects native
artifacts directly. The two are nonetheless separate implementations that
differ in many details; we list those most relevant to the results. We run
RewardKit~0.1.2 with its Codex judge (Codex~0.160.1) and the same OpenAI models
as Gandalf. Gandalf~1.0.0 uses the model-agnostic OpenHands SDK, splits each
\benchmarkabbrev rubric across two judge sessions, and writes a structured
verdict with criterion-specific reasoning and evidence; RewardKit judges the
full rubric in one session.

\newcommand{\rewardkitmatchedrows}{%
GPT-5.4 & 0.880 & 0.892 & $+0.012$ [$+0.004$, $+0.020$] & 34.75 & 49.46 \\
GPT-5.4-Mini & 0.884 & 0.883 & $-0.001$ [$-0.012$, $+0.010$] & 10.68 & 14.00 \\
GPT-5.4-Nano & 0.849 & 0.847 & $-0.001$ [$-0.021$, $+0.017$] & 4.27 & 7.58 \\
GPT-6.1-Sol & 0.854 & 0.849 & $-0.005$ [$-0.022$, $+0.010$] & 12.02 & 19.83 \\
}

\begin{table}[ht]
\caption{Matched-model comparison of RewardKit (RK) and Gandalf on all 24
\benchmarkabbrev tasks, pooling three evaluations per task. $\Delta$ is
Gandalf minus RewardKit with a nominal 95\% paired task-bootstrap interval
(10,000 draws; uncorrected for multiple comparisons). Cost is mean model API
cost per complete dataset pass. Both systems receive the task instruction,
reviewed workspace and trajectory, MCP access, and domain guidance; for
RewardKit, the guidance is a shortened version of Gandalf's, included in our
\href{https://github.com/Handshake-AI-Research/bankerverifierbench/blob/main/adapters/bvb/template/_tests-rewardkit/judge-prompt.md}{judge prompt}.}
\label{tab:rewardkit-matched}
\centering
\small
\setlength{\tabcolsep}{6pt}
\renewcommand{\arraystretch}{1.15}
\begin{tabular}{lrrcrr}
\toprule
Model & RK F1 & Gandalf F1 & $\Delta$ F1 [95\% CI] & RK cost (\$) & Gandalf cost (\$) \\
\midrule
\rewardkitmatchedrows
\bottomrule
\end{tabular}
\end{table}

Only the GPT-5.4 pair resolves an F1 difference, favoring Gandalf by
0.012; the Mini, Nano, and Sol pairs are unresolved. RewardKit costs
24--44\% less per pass in all four pairs. These are whole-system comparisons:
matching the model and evidence does not equate the harness-specific prompts,
interaction loop, or verdict format.

\newcommand{\rewardkitconditionrows}{%
RewardKit default inputs & 0.885 & 0.874 & 0.896 & 8.48 \\
RewardKit + matched evidence & 0.884 & 0.873 & 0.895 & 9.26 \\
RewardKit + matched evidence + guidance$^{\dagger}$ & 0.884 & 0.870 & 0.899 & 10.68 \\
Gandalf + matched evidence & 0.885 & 0.894 & 0.876 & 15.99 \\
Gandalf + matched evidence + guidance$^{\dagger}$ & 0.883 & 0.872 & 0.895 & 14.00 \\
}

\begin{table}[ht]
\caption{Input-condition comparison on \benchmarkabbrev with GPT-5.4-Mini at
high reasoning effort, evaluating all 24 tasks three times. Default inputs are
RewardKit's default prompt and workspace view, with the Codex judge and model
held fixed. Cost is per complete dataset pass. $^{\dagger}$The GPT-5.4-Mini
row of \Cref{tab:rewardkit-matched}.}
\label{tab:rewardkit-inputs}
\centering
\small
\setlength{\tabcolsep}{5pt}
\renewcommand{\arraystretch}{1.12}
\begin{tabular}{>{\raggedright\arraybackslash}p{7.2cm}rrrr}
\toprule
Condition & F1 & Precision & Recall & Cost (\$) \\
\midrule
\rewardkitconditionrows
\bottomrule
\end{tabular}
\end{table}
\FloatBarrier

No F1 difference in \Cref{tab:rewardkit-inputs} is statistically resolved.
RewardKit with its default inputs already reaches 0.885 F1, and adding the
task instruction, trajectory, MCP access, and guidance leaves it unchanged; at
matched evidence, Gandalf and RewardKit differ by 0.001~F1 (95\% CI
$[-0.013,0.013]$). This is consistent with both verifiers deriving most of their
accuracy from the shared reactive, in-environment approach; evidence and
guidance mainly change cost and, for Gandalf, the balance of precision and
recall. Guidance lowers Gandalf's precision by 0.023 ($[-0.043,-0.002]$) and
raises its recall by 0.019 ($[0.005,0.035]$); at matched evidence, Gandalf's
precision is 0.021 higher than RewardKit's ($[0.004,0.039]$).

On OpenClaw with GPT-5.4-Mini, RewardKit reaches 0.941 F1 at \$3.06 per pass
versus Gandalf's 0.923 at \$6.51; RewardKit's 0.018~F1 lead is not
statistically resolved (95\% CI $[-0.007,0.049]$). Inputs are not matched here:
RewardKit receives the real task instruction, full trajectory, workspace,
application-state tools, and a fixed OpenClaw judge prompt we wrote, whereas
Gandalf receives a generic verifier instruction, the final agent message,
workspace, and application-state tools, with no pointer to its trajectory copy
(\Cref{tab:gandalf-config}). We therefore treat this as a descriptive
cross-domain result rather than an architectural comparison.

\paragraph{Where the cost difference comes from.}
On \benchmarkabbrev with GPT-5.4-Mini, Gandalf costs \$0.583 per task versus
\$0.445 for RewardKit, and 84\% of this gap is output tokens. The main driver
is the number of model requests: Gandalf issues 42\% more per task (43.3 vs.\
30.5). RewardKit runs the model through Codex, OpenAI's native agent harness,
judges the full rubric in one session, and batches several tool calls into one
request in 7\% of requests. Gandalf runs it through the model-agnostic
OpenHands SDK, splits the rubric across two sessions, and batches in 3\% of
requests. On OpenClaw, where Gandalf is configured with one session, the
pattern recurs (21.5 vs.\ 14.5 requests per episode; 18\% vs.\ 37\% of
requests batched), although the gap there is mostly input tokens. Isolating
the contribution of each difference requires a controlled harness comparison,
which we leave to future work.

\section{Over-acceptance and over-rejection}
\label{app:error-direction}

With UNMET as the positive class (\Cref{sec:evaluation}), a false positive is
an \emph{over-rejection}: the verifier marks unmet a criterion the human
judged met. A false negative is an \emph{over-acceptance}: the verifier
marks met a criterion the human judged unmet. F1 weights the two equally,
but their consequences differ. Over-acceptance credits deficient outputs: if
the verifier supplies rewards, a policy may be reinforced toward outputs the
rubric fails, and in benchmarking it inflates the scores of deficient
systems. Over-rejection withholds credit from correct outputs: it can slow
learning or deflate benchmark scores, but it does not reward failures. We do
not measure either effect here. \Cref{tab:confusion} reports both error
types for the four Gemini~3~Flash verifiers.

\newcommand{\confusionrows}{%
Gandalf & 716.3 & 111.3 & 114.7 & 1521.7 & 0.864 & 1.03 & 221.3 & 6.3 & 14.7 & 153.7 & 0.955 & 2.32 \\
\midrule
Archipelago & 732.3 & 156.0 & 98.7 & 1477.0 & 0.852 & 0.63 & 230.0 & 80.3 & 6.0 & 79.7 & 0.842 & 0.07 \\
AAAJ & 535.0 & 159.0 & 296.0 & 1474.0 & 0.702 & 1.86 & 187.3 & 29.7 & 48.7 & 130.3 & 0.827 & 1.64 \\
AutoRubric & 686.3 & 130.0 & 144.7 & 1503.0 & 0.833 & 1.11 & 213.0 & 7.0 & 23.0 & 153.0 & 0.934 & 3.29 \\
}

\begin{table*}[t]
\caption{Confusion counts with UNMET as the positive class, as means per
complete dataset pass over three passes (Gemini~3~Flash). FP is
over-rejection (verifier unmet, human met); FN is over-acceptance (verifier
met, human unmet). \benchmarkabbrev has 2{,}464 criteria (33.7\% unmet by
human judgment) and OpenClaw 396 (59.6\% unmet); because counts and FN/FP
depend on this prevalence, they are not comparable across corpora.}
\label{tab:confusion}
\centering
\small
\setlength{\tabcolsep}{4pt}
\renewcommand{\arraystretch}{1.12}
\begin{tabular}{lrrrrrrrrrrrr}
\toprule
& \multicolumn{6}{c}{\benchmarkabbrev} & \multicolumn{6}{c}{OpenClaw} \\
\cmidrule(lr){2-7}\cmidrule(lr){8-13}
Verifier & TP & FP & FN & TN & F1 & FN/FP & TP & FP & FN & TN & F1 & FN/FP \\
\midrule
\confusionrows
\bottomrule
\end{tabular}
\end{table*}

We compare each baseline with Gandalf using nominal 95\% paired
task-bootstrap intervals for the per-pass difference (10{,}000 draws;
uncorrected for multiple comparisons). We report F1 intervals for context
but make directional claims only when an interval excludes zero. Verifiers
with similar F1 can differ in error direction. On \benchmarkabbrev,
Archipelago's F1 difference from Gandalf is
unresolved ($-0.012$, $[-0.034,0.006]$), but it makes 44.7 more
over-rejections per pass ($[20.0,67.3]$); its over-acceptance difference is
unresolved. AAAJ makes 181.3 more over-acceptances per pass
($[122.7,247.7]$), with its over-rejection difference unresolved. On
OpenClaw, Archipelago makes 74.0 more over-rejections per pass
($[47.7,102.0]$), and AAAJ makes more of both (over-acceptance $+34.0$,
$[13.3,64.7]$; over-rejection $+23.3$, $[6.0,48.3]$). Neither error-type
difference between AutoRubric and Gandalf is resolved on either corpus.

The backing model also shifts error direction. On \benchmarkabbrev, Gandalf
with GPT-6.1-Sol instead of Gemini~3~Flash has an unresolved F1 difference
($-0.015$, $[-0.044,0.013]$) but makes 92.3 fewer over-acceptances
($[-116.0,-71.0]$) and 155.0 more over-rejections ($[109.0,200.3]$) per
pass, lowering FN/FP from 1.03 to 0.08. GPT-5.4 shifts Gandalf the same
way (over-acceptance $-77.7$, $[-101.7,-55.3]$; over-rejection $+44.3$,
$[22.3,66.3]$), as it does Archipelago ($-26.7$, $[-50.3,-2.7]$; $+237.7$,
$[196.7,278.7]$). These observations are specific to these corpora and
configurations; they suggest reporting over-acceptance and over-rejection
alongside F1 when choosing a verifier.

\section{Prompts}
\label{app:taxonomy-prompts}
We provide the four prompts used by the pipeline below. Variables in
\verb|{{ ... }}| are populated at runtime with task and criterion
data.

\begin{tcolorbox}[
  colback=black!3!white,
  colframe=black!60!black,
  title={\textbf{Stage 1 prompt: capability discovery.}
Run once per criterion to emit free-text capability labels and a
verifier-side rationale.},
  boxrule=0.5mm,
  arc=3mm,
  breakable,
]
\small
\begin{Verbatim}[breaklines=true,breakanywhere=true,breaksymbolleft={},breaksymbolright={},baselinestretch=0.95]
You are a methodologist designing a capability taxonomy for evaluating
agentic/LLM-as-Judge verifiers (NOT for evaluating the agent). For each
rubric criterion below, name 1-3 short skills the VERIFIER must possess
to grade that criterion correctly, regardless of how the verifier is
implemented (tool-using judge, code-execution judge, multi-agent judge,
single-prompt judge, etc.).
 
# Verifier-side framing (read carefully)
 
A "capability" here is a property of the VERIFIER, not of the agent.
 
- Right framing: "the verifier needs to <verb a verifier does> <noun
  describing what it inspects in the deliverable or task spec>". Example
  verbs: inspect, locate, recompute, cross-check, recognize, decompose,
  trace, measure, compare, audit, parse, interpret, classify.
- Wrong framing (DO NOT produce): naming what the agent has to do to
  satisfy the criterion. Example forbidden agent-construction head-verbs:
  build, construct, produce, create, design, develop, write, draft,
  generate, populate, deliver, prepare, assemble, compile, render.
  (Note: words like `document`, `layout`, `output`, `format`, `present`
  are FINE when used as nouns or adjectives, e.g.
  `document_title_content_verification_capability` or
  `present_value_audit_capability`.)
- Wrong framing (DO NOT produce): naming a specific tool, file format,
  or runtime. Forbidden tokens anywhere in the capability name: excel,
  xlsx, pptx, pdf, openpyxl, pandas, python, jinja, json, csv, mcp,
  sec, edgar, sql, api.
- The verifier is judge-architecture-agnostic. The capability MUST be
  expressible for a verifier that:
  (a) reads the deliverable file and computes things itself,
  (b) reads the agent's generation script as proxy,
  (c) calls a remote tool to fetch source-of-truth data,
  (d) reasons over a rendered preview without executing code.
  If your capability only makes sense for one of these, restate it more
  abstractly.
 
# Output format (HARD constraints)
 
Each capability name MUST satisfy ALL of:
 
- 4 to 8 words inclusive.
- lower_snake_case (only [a-z_]; no digits, no hyphens, no spaces).
- ENDS WITH `_capability` (counts as one of the 4-8 words).
- Reads as "<verb-or-adjective>_<thing-the-verifier-inspects>_capability".
- Distinct from sibling capabilities WITHIN this same rubric category.
  If two criteria in the same category map to the same capability, only
  emit it once per criterion, but make sure the contrasting one (if any)
  is genuinely different, not a paraphrase.
 
Good examples (illustrative archetypes -- DO NOT just copy them; produce
labels that fit the actual criterion):
 
- `numeric_value_recomputation_capability`
- `formula_dependency_tracing_capability`
- `cross_section_consistency_checking_capability`
- `external_source_factual_lookup_capability`
- `unit_and_scale_normalization_capability`
- `qualitative_intent_interpretation_capability`
- `multi_period_completeness_auditing_capability`
- `visual_layout_inspection_capability`
 
Bad examples (rejected by post-validation; do NOT produce these):
 
- `excel_workbook_check_capability` -> contains forbidden token `excel`.
- `build_dcf_model_capability` -> agent-side head-verb `build`.
- `formula_audit` -> missing `_capability` suffix.
- `verifier_reads_the_workbook_thoroughly_to_validate_capability` -> >8 words.
- `recompute_capability` -> <4 words; not specific.
- `accuracy_capability` -> generic virtue, not a verifier skill.
 
# Each criterion's rationale (also required)
 
For each criterion, include a `rationale` string of 1-2 sentences,
explaining why the verifier needs the capabilities you listed. The
rationale MUST:
 
- Start literally with "the verifier needs to ".
- Mention something concrete from the criterion text (a quoted phrase,
  a number, a section name) so we can audit grounding.
- NOT mention the agent's process or how the agent should accomplish
  anything.
 
# Task context (read-only)
 
Task ID: {{ task_id }}
Cohort: {{ cohort }}
Rubric category: {{ rubric_category }}
 
## Task brief (instructions the agent received)
 
This is provided ONLY so you can understand WHAT deliverable the
verifier will inspect. The capability you name is about INSPECTING the
deliverable, not about the agent's process to produce it.
 
<task_brief>
{{ task_brief }}
</task_brief>
 
# Criteria to annotate ({{ n }} criterion(a) in this batch)
 
{% for c in criteria %}
### Criterion {{ loop.index0 }}
- **Text**: {{ c.criterion }}
- **Category**: {{ c.category }}
- **Direction (FYI only, do not let this bias you)**: {{ c.direction }}
 
{% endfor %}
 
# Output
 
Return a JSON array, one element per criterion, in the SAME order as
above.
 
[
  {
    "criterion_idx": 0,
    "capabilities": ["<lower_snake_case_capability_1>", "<...>"],
    "rationale": "the verifier needs to ..."
  }
]
 
Constraints again, summarised:
 
1. 1-3 capabilities per criterion (never 0, never >3).
2. Each capability: 4-8 words, lower_snake_case, MUST end with
   `_capability`.
3. No agent-side head-verbs (build/produce/create/design/develop/...).
4. No tool/format tokens (excel/xlsx/python/openpyxl/pandas/json/csv/
   sec/edgar/api/sql).
5. Rationale starts with "the verifier needs to " and quotes from the
   criterion.
6. Output ONLY the JSON array. No preamble, no commentary, no markdown
   fence.
\end{Verbatim}
\normalsize
\end{tcolorbox}

\begin{tcolorbox}[
  colback=black!3!white,
  colframe=black!60!black,
  title={\textbf{Stage 2 prompt (a): per-cluster canonicalization.} Run once per micro-cluster to assign a canonical name, definition,
diagnostic signal, example, and contrastive non-examples.},
  boxrule=0.5mm,
  arc=3mm,
  breakable,
]
\small
\begin{verbatim}
You are a methodologist consolidating short, machine-tractable
verifier-capability labels for an agentic/LLM-as-Judge evaluation
taxonomy.
 
# What is a "capability"?
 
A capability is a property of the VERIFIER (the
LLM-as-Judge/Agent-as-a-Judge/Any-other-judge), not of the agent. It names a
verification skill the verifier needs in order to grade a rubric
criterion correctly, regardless of whether the verifier uses tools,
executes code, or just reasons over a rendered preview. Capability
labels MUST NOT mention specific tools, file formats, or runtimes.
 
# Input
 
Below is a single CLUSTER of {{ n_members }} member-labels emitted by a
Phase 1 free-text discovery pass. They have been grouped by a
deterministic embedding + clustering step. Most members are
paraphrases of the same underlying verifier skill; some members may be
drift.
 
Cluster id: `{{ cluster_id }}`
Member-label frequency in the corpus (ranked):
 
{% for m in member_labels %}- `{{ m.label }}`  (frequency: {{ m.freq }})
{% endfor %}
 
A few representative example criteria from this cluster (verbatim from
the rubric):
 
{% for ex in example_criteria %}- *(task: {{ ex.task_id }}, category:
{{ ex.category }})* {{ ex.criterion }}
{% endfor %}
 
# Task
 
Pick ONE canonical name for this cluster, plus a short description, a
diagnostic signal, an example pattern, and 1-3 contrastive non-examples
("if it's about X, prefer the `<sibling>` capability instead").
 
## Naming constraints (HARD)
 
The canonical `name` MUST satisfy ALL of:
 
- 4 to 8 underscore-separated words (counting `_capability` as one
  word).
- lower_snake_case, only [a-z_].
- Ends with `_capability`.
- Reads as "<verb-or-adjective>_<thing-the-verifier-inspects>_capability".
- No agent-side construction verbs (build/construct/produce/create/
  design/develop/write/draft/generate/populate/deliver/prepare/
  assemble/compile/output/render).
- No tool/format tokens (excel/xlsx/pptx/pdf/openpyxl/pandas/python/
  jinja/json/csv/mcp/sec/edgar/sql/api).
- Distinct enough from common siblings 
(cross-section consistency, external lookup, completeness/
  scope, semantic intent, unit normalization, structural-grading) that
  it could not be paraphrased into one of them.
 
## Output
 
Return a single JSON object (NOT an array):
 
{
  "name": "<lower_snake_case_capability>",
  "description": "<1-2 sentences: what observable verification skill
    this is>",
  "diagnostic_signal": "<1 sentence: what is visible in the trace if
    the verifier exercised this capability vs. failed to>",
  "example_pattern": "<1-2 sentences: a concrete narrative of the
    verifier exercising this skill on a deliverable>",
  "non_examples": [
    "if the criterion is about <adjacent skill X>, prefer the
    `<sibling_capability>` label.",
    "..."
  ]
}
 
No preamble, no commentary, no markdown fence -- only the JSON object.
\end{verbatim}
\normalsize
\end{tcolorbox}

\begin{tcolorbox}[
  colback=black!3!white,
  colframe=black!60!black,
  title={\textbf{Stage 2 prompt (b): final merge.} Run once over all canonicalized clusters to produce the final
taxonomy of 8--10 capabilities under the size and coverage
constraints.},
  boxrule=0.5mm,
  arc=3mm,
  breakable,
]
\small
\begin{verbatim}
You are finalising a small, distinct verifier-capability taxonomy 
that will help characterize what a verifier running in 
a Reinforcement Learning evironment needs to do. 
You are given a slate of {{ n_in }}
candidate capabilities already produced by a clustering step, and you
must merge near-duplicates and produce a final list of {{ min_caps }}-
{{ max_caps }} canonical capabilities.
 
# Background
 
A capability is a property of the VERIFIER, not of
the agent. Each canonical capability must be expressible for verifiers
of any architecture (tool-using, code-executing, or pure-reasoning).
Capability names MUST NOT mention specific tools, file formats, or
runtimes.
 
# Candidate slate
 
Each candidate has its current name, description, and the share of
corpus criteria (out of {{ n_total }}) where it was the dominant tag at
the cluster level.
 
{% for c in candidates %}- name: `{{ c.name }}`
  share: {{ c.share }}%  (n_criteria_dominated: {{ c.n }})
  description: {{ c.description }}
  diagnostic_signal: {{ c.diagnostic_signal }}
{% endfor %}
 
# Merge rules (HARD)
 
1. **Final count.** Output between {{ min_caps }} and {{ max_caps }}
   canonical capabilities (inclusive).
2. **No giant bucket.** No single canonical capability may dominate
   more than {{ max_share_pct }}% of the corpus. If a candidate
   currently exceeds this share, propose a SPLIT: emit two or more
   sub-capabilities whose union covers the same members but each
   respects the cap.
3. **No singletons.** No canonical capability may dominate fewer than
   2% of the corpus. Merge such candidates into the nearest semantic
   neighbor.
4. **Pairwise distinctness.** For every pair of canonical capabilities
   (A, B) that could plausibly be confused, A's `non_examples` or B's
   `non_examples` MUST mention the other and explain the boundary.
5. **Coverage.** Every input candidate must map to exactly one
   canonical capability via the `member_input_names` field below. No
   silent drops.
 
# Naming constraints (HARD)
 
Same as in the upstream prompt:
 
- 4 to 8 underscore-separated words (counting `_capability` as one
  word).
- lower_snake_case, only [a-z_].
- Ends with `_capability`.
- Reads as "<verb-or-adjective>_<thing-the-verifier-inspects>_capability".
- No agent-side head-verbs (build/produce/create/design/develop/write/
  draft/generate/populate/deliver/prepare/assemble/compile/output/
  render).
- No tool/format tokens (excel/xlsx/pptx/pdf/openpyxl/pandas/python/
  jinja/json/csv/mcp/sec/edgar/sql/api).
 
# Output
 
Return a JSON array:
 
[
  {
    "name": "<lower_snake_case_capability>",
    "description": "<1-2 sentences>",
    "diagnostic_signal": "<1 sentence>",
    "example_pattern": "<1-2 sentence narrative>",
    "non_examples": ["if <pattern>, prefer `<sibling_capability>`",
      "..."],
    "member_input_names": ["<candidate name 1>", "<candidate name 2>",
      "..."]
  }
]
 
No preamble, no commentary, no markdown fence -- only the JSON array.
\end{verbatim}
\normalsize
\end{tcolorbox}

\begin{tcolorbox}[
  colback=black!3!white,
  colframe=black!60!black,
  title={\textbf{Stage 3 prompt: closed-vocabulary tagging.} Run once per criterion to assign one to three taxonomy capabilities,
ranked by confidence.},
  boxrule=0.5mm,
  arc=3mm,
  breakable,
]
\small
\begin{verbatim}
You are a methodologist tagging rubric criteria with the verifier
capabilities required to grade them. The taxonomy below is fixed. Pick
1-3 capabilities per criterion from the closed vocabulary -- do NOT
invent new names.
 
# Verifier-side framing (read carefully)
 
A "capability" is a property of the VERIFIER, not of
the agent. It names a verification skill the verifier must possess to
grade the criterion correctly, regardless of the verifier's internal
architecture (tool-using, code-executing, single-prompt, multi-agent,
etc.).
 
This is judge-architecture-agnostic on purpose: the same tag applies
whether the verifier reads the deliverable file directly, parses the
agent's generation script, calls a remote tool, or reasons over a
rendered preview.
 
# Capability taxonomy (closed vocabulary -- pick from these only)
 
{% for cap in capabilities %}
### `{{ cap.name }}`
- **Description**: {{ cap.description }}
- **Diagnostic signal**: {{ cap.diagnostic_signal }}
- **Example pattern**: {{ cap.example_pattern }}
{% if cap.non_examples %}- **Boundaries**:
{% for ne in cap.non_examples %}    - {{ ne }}
{% endfor %}{% endif %}
{% endfor %}
 
# Task context
 
Task ID: {{ task_id }}
Cohort: {{ cohort }}
Rubric category: {{ rubric_category }}
 
## Task brief (instructions the agent received)
 
This is provided ONLY so you can understand what deliverable the
verifier will inspect. The capability you assign is about INSPECTING
the deliverable, not about how the agent might produce it.
 
<task_brief>
{{ task_brief }}
</task_brief>
 
# Criteria to tag ({{ n }} criteria in this batch)
 
{% for c in criteria %}
### Criterion {{ loop.index0 }}
- **Text**: {{ c.criterion }}
- **Category**: {{ c.category }}
 
{% endfor %}
 
# Tagging rules
 
1. Each criterion MUST be assigned 1-3 capability names, in
   MOST-RELEVANT-FIRST order.
2. Capability names MUST exactly match a `name` from the taxonomy above
   (case-sensitive, including the `_capability` suffix). No paraphrasing.
3. Tag based on what the criterion REQUIRES the verifier to do, not on
   whether any specific verifier implementation actually does it.
4. Reserve very-broad / structural-grading capabilities for criteria
   where the grading process itself is the focus (e.g. a self-consistency
   check), not as a default top-up tag.
 
# Output
 
Return a JSON array, one element per criterion, in the SAME order as
above. Each element MUST set `criterion_idx` to the criterion's index
above.
 
[
  {
    "criterion_idx": 0,
    "capabilities": ["<capability_name_1>", "<capability_name_2>",
      "..."]
  }
]
 
Output ONLY the JSON array. No preamble, no commentary, no markdown
fence.
\end{verbatim}
\normalsize
\end{tcolorbox}

\end{document}